\documentclass[12pt]{article}
\usepackage{mathtools}   
\usepackage[linesnumbered,algoruled,boxed,ruled,vlined]{algorithm2e}
\SetAlCapSkip{1em} 
\SetKwInput{KwParam}{Parameters} 
\SetKwInput{KwHyper}{Hyperparameters} 
 
\SetCommentSty{mycommfont} 
\usepackage{amsthm}
\usepackage{algorithmicx}
\usepackage{algpseudocode}
\usepackage{xcolor}
\usepackage{amsfonts}  
\usepackage{amsmath}   
\usepackage[linesnumbered,algoruled,boxed,ruled,vlined]{algorithm2e}
\usepackage{float}
\begin{document}
\baselineskip 12pt
\linespread{2} 
 \begin{center}
\textbf{Enhancing Object Detection in Ancient Documents with Synthetic Data Generation and Transformer-Based Models} \\

\vspace{1.5cc}
{ \sc Zahra Ziran$^{1}$, Francesco Leotta$^{1}$, Massimo Mecella$^{1}$}\\

\vspace{0.3 cm}

{\small $^{1}$Sapienza University of Rome}\\
\textit{\{name.surname\}@uniroma1.it}

 \end{center}

\vspace{1.5cc}

\begin{abstract}
  \noindent  The study of ancient documents provides a glimpse into our past. However, the low image quality and intricate details commonly found in these documents present significant challenges for accurate object detection. The objective of this research is to enhance object detection in ancient documents by reducing false positives and improving precision. To achieve this, we propose a method that involves the creation of synthetic datasets through computational mediation, along with the integration of visual feature extraction into the object detection process. Our approach includes associating objects with their component parts and introducing a visual feature map to enable the model to discern between different symbols and document elements. Through our experiments, we demonstrate that improved object detection has a profound impact on the field of Paleography, enabling in-depth analysis and fostering a greater understanding of these valuable historical artifacts.

\vspace{0.95cc}
\parbox{24cc}{{\it Document Image Analysis, Ancient Document, Feature Extraction and Transformer Model}:
}
\end{abstract}

\section{Introduction} \label{introduction}
The study of ancient documents is crucial for understanding the past and making these valuable resources accessible to a wider audience. Accurate object detection in these documents is essential for deciphering their content and context \cite{ref-notae, ref-symbol-detection}. However, the unique characteristics of ancient documents, such as faded or damaged text, non-standardized writing systems, and complex layouts, make object detection a challenging task. In this paper, we propose a novel method for improving object detection in ancient documents by leveraging synthetic data generation, transformer-based models, and tokenization techniques \cite{ref-statistics}.
Our proposed method involves extracting visual features through tokenization and defining an optimization task to fit a transformer model on graphic symbol data for the image classification task. This approach is designed to capture the unique characteristics of ancient documents and enhance classification accuracy. We then use the image classifier with a fast search method to gather object detection annotations and train a Faster R-CNN model \cite{Fast r-cnn,ref-detectors}.
A critical aspect of our method is the generation of an identical synthetic dataset, which is essential for accurate object detection. The resulting dataset is paired with the original data, creating a dictionary-like relationship. This relationship is useful for approximating continuous functions that transform vectors from the vector space of objects into the vector space of geometrical trapezoids, thus preserving the inherent structure and characteristics of the ancient documents. We present a symbol-level tokenization scheme, where $\textit{X}$ is a binary image that is transformed by replacing each element with a tuple of its pixel index in the image.
By integrating synthetic data generation, transformer-based models \cite{ref-transformer,ref-algorithms}, and tokenization techniques, our method aims to enhance object detection in ancient documents, by learning the map between representations at part-level and object-level, \cite{ref-capsules}. In the section \ref{sec:methods}, we  provide a pseudo-code algorithm to demonstrate the feature extraction process and showcase how our method is implemented.

\section{Materials and Methods} \label{sec:methods} 
The proposed method consists of a multi-step algorithm for detecting and classifying graphical symbols in images. In the first step, the input image is preprocessed to remove noise and convert it to a binary format. This is achieved by applying a series of image processing techniques \cite{ref-scikit}, including thresholding and morphological operations. Next, the image is segmented into stroke segments, which are defined as contiguous regions of black pixels in the binary image. The coordinates of these stroke segments are then clustered using the OPTICS algorithm \cite{ref-optics}, which automatically identifies spatially related groups of stroke segments. A centroid is computed for each cluster by taking the average of the coordinates of the points in the group.

To create variable stroke thickness, two circles with variable radii are constructed at each end of the stroke segment. For each cluster, a circle is defined by calculating the centroid and its associated radius. The radius is determined by computing the mode of an array of Euclidean distances between the centroid and all other points in the cluster. The centroids and radii are then used to construct trapezoids that represent the strokes of the graphical symbols. The trapezoid vertices are defined as the intersecting points of the trapezoid sides with the perimeter outlines of the centroids. The trapezoids' vertex labels are oriented in a clockwise manner so that trapezoids from all clusters have the same orientation relative to the image area. The density of each trapezoid is computed by counting the foreground pixel values that are trapped inside the shape and dividing the sum by the area of the trapezoid. The resulting feature vector of a trapezoid includes four vertices and one brightness density related to the shape. The location of a bounding box is fully described by a set of affine transformations, which are subject to four axioms: closure, associativity, identity, and inverse. The affine transformations are used to approximate the transform \texttt{T} that maps the original image to the bounding boxes of the graphical symbols. The inverse of transform \texttt{T} is used to build a database of paired instances for producing synthetic training data \cite{ref-gan}\cite{ref-omniglot}.
\subsection{step-by-step pseudo-code algorithm}
In this section, we provide a step-by-step pseudo-code algorithm for our proposed method for feature extraction in ancient documents.
\begin{enumerate}
    \item Perform random sampling on the binary image, obtaining an array of coordinates. 

    \item Cluster the samples from step 1 using the OPTICS algorithm with minimum points set to 5 and epsilon set to infinity. Calculate the centroids for each cluster by taking the average of the coordinates of the points in the group. 

    \item Connect the calculated centroids to construct trapezoids.

    \item For each cluster i, create a circle $c_i$ centered on the cluster core point and with a radius $r_i$ equal to the mode of the Euclidean distances from $c_i$ to all other points in cluster i.

    \item Ignore noise points, which are not inside any centroid (negative labels).

    \item Repeat steps 4 and 5 for cluster $i$, where $i$ runs over from $1$ till all clusters are iterated.
    \item Calculate the radii of the centroids using the mode of an array of distances to the centroid.

    \item Transform the image of the symbol into the number of clusters ($N_C$).

    \item Link centroids $c_i$ and $c_j$ with a straight line and draw a pair of trapezoid sides, each perpendicular to the connecting segment, one containing centroid $c_i$ and the other containing centroid $c_j$.

    \item Define the intersecting points of the trapezoid sides with the perimeter outlines of the centroids as vertices $P^1$, $P^2$, $P^3$, and $P^4$.

    \item Label the vertices such that the segment connecting $P^1$ to $P^2$ measures 2 × $r_i$ in length and the segment connecting $P^3$ to $P^4$ measures 2 × $r_j$, where $r_j$ and $r_j$ are the corresponding radii related to centroids $c_i$ and $c_j$, respectively.

    \item Rename the trapezoid vertices in a clockwise manner to ensure a consistent reference frame.

    \item Connect every pair of centroids $c_i$ and $c_j$ with a unique trapezoid.

    \item Trace each centroid back to the original image and calculate the density of each centroid by dividing the sum of the foreground pixel values by the area of the circle ($\pi \times r^2$) for a centroid, or by the area of the trapezoid (($r_i + r_j) \times ||c_i - c_j||^2$) for a trapezoid.

    \item Create a feature vector for each trapezoid, including its four vertices and the brightness density related to the shape.

    \item Filter noise using the calculated trapezoid surface area as the basis.

\end{enumerate}

By following these steps, the proposed method extracts feature from ancient documents, enabling accurate object detection and classification. This pseudo-code algorithm serves as a guideline for implementing the method in a scientific paper or research project.


\begin{algorithm}[H] \label{ref-algorithm}
{
    \DontPrintSemicolon 
    \SetAlgoLined
    
    \caption{$\mathbf{Y} \leftarrow \texttt{T}(\mathbf{X} | N_T, d_x)$}
    
    
    \KwIn{$\mathbf{X} \in \mathbb{R}^{d_x \times d_x}$, vector representation of graphic symbols.}
    
    \KwOut{$\mathbf{Y} \in \mathbb{R}^{d_x \times d_x}$, updated representations of symbols in $\mathbf{X}$, folding in information from their geometric shapes.}
    
    \KwHyper{$N_T \in \mathbb{N}$, max number of trapezoids to visualize, $d_x$, image side length.}

    $E \leftarrow \{(i, j) | i, j \in \{1, 2, ..., d_x\}, ~ \forall ~ i, j, ~ \mathbf{X}[i, j]\}$
    
    \tcp{\footnotesize Estimate clustering structure from vector array. $cluster: \mathbb{N}^{d_x ~ \times ~ d_x ~ \times ~ 2} \longrightarrow \mathbb{R}^{N_C ~ \times N_{...} ~ \times ~ 2}$.}
    
    $C \leftarrow cluster(E)$

    $N_C \leftarrow length(C[:, 1, 1])$   \hspace{1cm}\tcc{Count the number of clusters}

    \For{$i = 1, 2, ..., N_C$}
    {

        $N_i \leftarrow length(C[i,:,1])$    \hspace{1cm}\tcc{Count the points in cluster i}

        $c_i \leftarrow \frac{1}{N_i} \sum_{k=1}^{N_i} C[i,k,:]$ \hspace{1cm}\tcc{Calculate the center of cluster i}

        \tcc{Calculate the radius of cluster j}
        
        $r_i \leftarrow mode(magnitude([C[i, k,:] - c_i)) ~ for ~ k \in \{1, 2, ..., N_i\}$

            
            \For{$j = 1, 2, ..., N_C$}
            {

                $N_j \leftarrow length(C[j,:,1])$   \hspace{1cm}\tcc{Count the points in cluster j}

                $c_j \leftarrow \frac{1}{N_j} \sum_{k=1}^{N_j} C[j,k,:]$ \hspace{1cm}\tcc{Calculate the center of cluster j}

                \tcc{Calculate the radius of cluster j}
        
                $r_j \leftarrow mode(magnitude([C[j, k,:] - c_j)) ~ for ~ k \in \{1, 2, ..., N_j\}$
                
                $P^1, P^2, P^3, P^4 \leftarrow construct\_trapezoid(c_i, r_i, c_j, r_j)$

                $M[i,j] \leftarrow count\_foreground\_pixels(\mathbf{X}, P^1[i,j], P^2[i,j], P^3[i,j], P^4[i,j])$ 

                $D[i,j] \leftarrow calculate\_area(P^1[i,j], P^2[i,j], P^3[i,j], P^4[i,j])$ 

                \tcc{Calculate brightness density for trapezoid associated to i and j.}
                $\mathbf{S}[i,j] \leftarrow \frac{M[i,j]}{D[i,j]}$ 
            } 
    }

    $\mathbf{s} \leftarrow flatten(\mathbf{S})$ 

    $\mathbf{p^1}, \mathbf{p^2}, \mathbf{p^3}, \mathbf{p^4} \leftarrow flatten(P^1), flatten(P^2), flatten(P^3), flatten(P^4)$

    $ sort((\mathbf{p^1}, \mathbf{p^2}, \mathbf{p^3}, \mathbf{p^4}, \mathbf{s}), criteria = \mathbf{s}, decreasingly = true)$

    \tcc{Draw $N_T$ trapezoids for a cleaner plot.}
    $\mathbf{Y} \leftarrow plot\_trapezoid(\mathbf{p^1}[k], \mathbf{p^2}[k], \mathbf{p^3}[k], \mathbf{p^4}, \mathbf{s}[k]) ~ for ~ k ~ \in ~ \{1, 2, ..., N_T\}$

	\Return $\mathbf{Y}$
 }
\end{algorithm}


\section{Conclusion} \label{sec:conclusion} 
The proposed method involves a multi-step algorithm for feature extraction, including preprocessing of the input image, segmentation of stroke segments, clustering of coordinates, and construction of trapezoids to represent graphical symbols. Synthetic data generation is used to create a paired dataset for accurate object detection. The method also incorporates transformer-based models and tokenization techniques to capture the unique characteristics of ancient documents. By following the step-by-step pseudo-code algorithm, researchers can implement the proposed method in their own projects and scientific papers. The algorithm provides guidance for extracting features from ancient documents and facilitating accurate object detection and classification. In conclusion, the presented method offers a valuable contribution to the field of document analysis by effectively tackling the difficulties caused by the low image quality and intricate details of ancient documents. Through the enhancement of object detection, researchers are empowered to delve deeper into the historical context and glean valuable insights from these invaluable resources. Moreover, the improved accessibility of ancient documents to a broader audience facilitates wider engagement and understanding of our past. To further expand the scope of its applicability, future research and experimentation can explore the potential of this method in various other domains of historical document analysis and digitization.




\end{document}